\documentclass[letterpaper, 10 pt, conference]{ieeeconf}  

\usepackage[cmex10]{amsmath}
\usepackage{amssymb}
\usepackage{mathrsfs}
\usepackage{graphicx}
\usepackage{float}
\usepackage{array}
\usepackage{epstopdf}
\usepackage{multirow}

\IEEEoverridecommandlockouts                              

\title{\LARGE \bf
A Deep Unsupervised Learning Approach Toward MTBI Identification Using Diffusion MRI}

\author{Shervin~Minaee$^{1}$,  Yao~Wang$^1$, Anna Choromanska$^1$, Sohae Chung$^2$, Xiuyuan Wang$^2$, Els Fieremans$^2$ \\
Steven Flanagan$^3$, Joseph Rath$^3$, Yvonne W. Lui$^2$
\thanks{*This work was supported by  National Institutes of Health (NIH) under award number  R21NS090349.}%
\thanks{$^{1}$Shervin Minaee, Yao Wang and Anna Choromanska are with the department of Electrical and Computer Engineering, New York University, New York, NY, 11201, USA
        {\tt\small (email:\ shervin.minaee@nyu.edu)}.} %
\thanks{$^{2}$Sohae Chung, Xiuyuan Wang, Els Fieremans, and Yvonne W. Lui are with the Department of Radiology, New York University
        New York, NY, 10016, USA.
       }
\thanks{$^{3}$Steven Flanagan and Joseph Rath are with the Department of Rehabilitation Medicine, New York University, New York, NY, 10016, USA.
       }
}

\begin{document}

\maketitle
\thispagestyle{empty}
\pagestyle{empty}

\begin{abstract}
Mild traumatic brain injury is a growing public health problem with an estimated incidence of over 1.7 million people annually in US. 
Diagnosis is based on clinical history and symptoms, and accurate, concrete measures of injury are lacking.
This work aims to directly use diffusion MR images obtained within one month of trauma to detect injury, by incorporating deep learning techniques.
To overcome the challenge due to limited training data,  we describe each brain region using the bag of word representation, which specifies  the distribution of representative patch patterns. 
We apply a convolutional auto-encoder to learn  the patch-level features, from overlapping image patches extracted from the MR images,  to learn features from  diffusion MR images of brain using an unsupervised approach.
Our experimental results show that the bag of word representation using  patch level features learnt by the auto encoder provides similar performance as that using the raw patch patterns, both significantly outperform earlier work relying on the mean values of MR metrics in selected brain regions.
\end{abstract}

\section{INTRODUCTION}
Mild traumatic brain injury (mTBI) is a growing public health problem, which can lead to a variety of problems including persistent headache, memory and attention deficits, as well as affective symptoms.
A building body of works show that diffusion MRI can reveal subtle brain injury; however, no single imaging metric has been shown to be sufficient as an independent biomarker \cite{lui3}-\cite{lui5}.

While diffusion MRI is extremely promising in the study of mTBI, a definitive method of diagnosing patients with recent mTBI remains a challenge.
Both gray matter such as the thalamus, and white matter such as the corpus callosum (CC) and frontal deep white matter have been repeatedly implicated as areas at high risk for injury in the literature.
While there have been a few previous studies using machine learning for mTBI identification from MR images \cite{yuanyi}-\cite{minaee2},  the features used in those works are mainly hand-crafted and may  not be the most discriminative features for this task.

In this work, we develop a machine learning framework to classify mTBI patients and controls using features extracted from diffusion MRI in the thalamus and corpus callosum, two regions highly implicated in imaging based on previous works \cite{thal}-\cite{cc}.
The main challenge for using a machine learning approach is that we only have limited samples (114 subjects), and each sample has a very high dimensional raw representation (multiple 3D volumes). Therefore, it is not possible to directly train a classification network using the raw MRI volume data as the input.
We  propose a new approach for feature extraction from MR images, where  we first learn the feature representation of patches using a deep unsupervised learning approach \cite{dnn}, and then aggregate the features from different patches through a bag of word representation, and use them along with demographic and neuro-cognitive test features as the overall feature vector.
We then use feature selection followed by a classification algorithm to identify mTBI patients.
The block diagram of the overall algorithm is shown in Fig. 1.
Through experimental study, we show that by learning patch level deep features and aggregating them through
a bag of word representation for each brain region, we can achieve much higher accuracy compared to using mean values of the various MR metrics in each region.
\begin{figure}[h]
\begin{center}
    \includegraphics [scale=0.42] {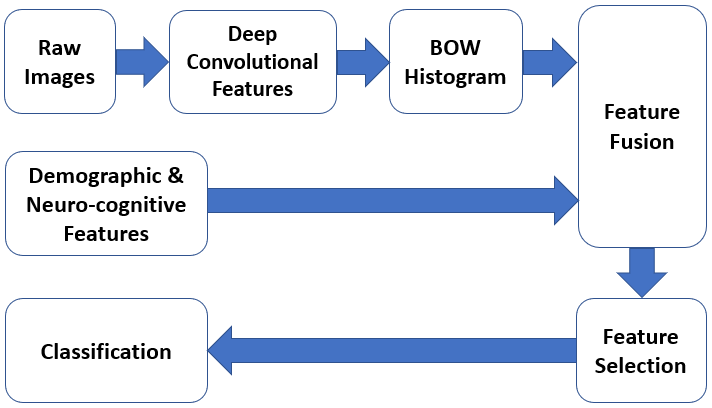}
\end{center}
  \caption{The block-diagram of the proposed mTBI identification algorithm}
\end{figure}


The rest of the paper is organized as follows. 
Section II provides the description of the proposed framework. 
Section III provides the experimental studies and comparison to other works. 
And finally the paper is concluded in Section IV.

\section{The Proposed Framework}
There have been some previous studies on mTBI classification using various sets of features, from demographic (such as age and gender) and neurocognitive  to hand-crafted features from medical images.
In this work we propose a framework for mTBI identification that incorporates all the imaging, demographics, and neuro-cognitive features.
We derive the imaging features from multi-shell diffusion MR imaging, and bicompartment modeling based on white matter tract integrity (WMTI) metrics derived from diffusion kurtosis imaging (DKI) \cite{els}, that are shown to be promising for assessment of mTBI patients against controls \cite{lui3}-\cite{lui5}.
These metrics are summarized in Table I.
\begin{table}[h]
\centering
\caption{MRI metrics description}
\begin{tabular}{| m{2cm} | m{5.5cm} |}
\hline
MRI Metric & Metric Description  \\
\hline
AWF & Axonal Water Fraction  \\
\hline
Da & Diffusivity within Axons  \\
\hline
De-par & Diffusion parallel to the axonal tracts in the extra-axonal  \\
\hline
De-perp & Diffusion perpendicular to the axonal tracts in the extra-axonal  \\
\hline
FA & Fractional Anisotropy  \\
\hline
MD & Mean Diffusion  \\
\hline
AK,\ MK,\ RK & Axial Kurtosis, Mean Kurtosis, Radial Kurtosis  \\
\hline
\end{tabular}
  \label{tab:1}
\end{table}


Because of the limitation of the number of samples, it is not possible to train a deep convolutional network to directly classify the brain images. 
To tackle this problem, and also based on the assumption that mTBI may impact only certain regions in the brain, we propose to represent each brain region by a bag of words (BoW) representation, which is the histogram of different clustered  patch (i.e. a small 2D slice in brain) patterns among all patches in the region. 

A main challenge in using such a representation is how to describe each local patch. 
Because mTBI does not necessarily impact all patches, we cannot infer the patch-level  labels from subject-level labels. 
Therefore we cannot learn patch-level features through supervised learning. To overcome this problem, we apply an unsupervised learning approach at the patch level, and train an auto-encoder to learn features that can be used to reconstruct patches.
In the following,  we first explain how we learn patch level features through training an auto-encoder, and then describe how we aggregate the patch-level features using a bag of words representation for each region, and finally how to combine the region representations and other information for patient-level classification.

\subsection{Patch Feature Learning Using an Auto-Encoder}
In order to learn patch-level features without having labels, we employ a convolutional auto-encoder \cite{CAE}-\cite{cae2},  an unsupervised feature learning approach.
Convolutional auto-encoder receives an image patch as the input and performs multiple convolution plus downsampling layers to encode the image into some latent representation, and then uses these features to reconstruct the original patch  through deconvolution.
By doing so,  the network is forced to learn some representative information that is sufficient to recover the image.
The overall architecture of a convolutional auto-encoder is shown in Figure 2.
Here, each layer of the network performs three operations: convolution, nonlinear transformation, and pooling (downsampling).
After training this model, the latent representation in the mid-layer is used as patch feature representation.
\begin{figure}[h]
\begin{center}
    \includegraphics [scale=0.34] {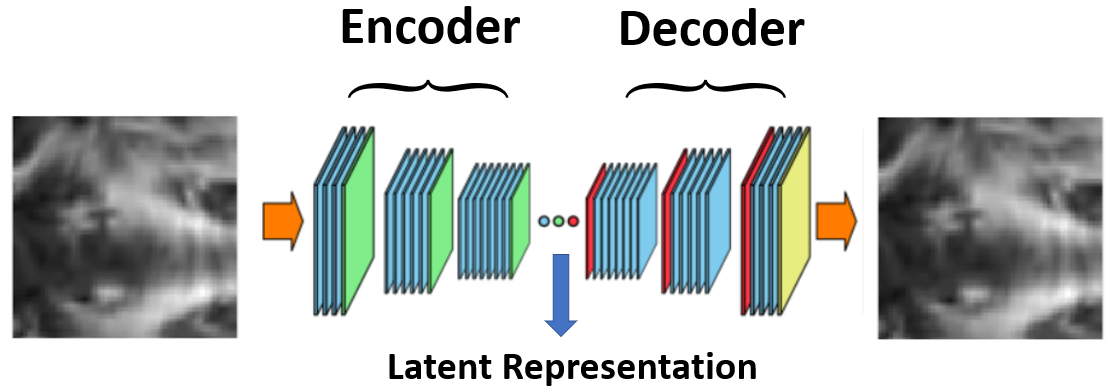}
\end{center}
  \caption{The block-diagram of the proposed convolutional auto-encoder}
\end{figure}

In our study we consider two scenarios when training the convolutional auto-encoder.
In one scenario, we train one model for each metric (such as FA, MK, RK, etc.).
In the other scenario, we concatenate all metrics together (treat them as different channels) to form a 3D patch and train one convolutional auto-encoder.
We use the same model for all regions (Thalamus and CC).

\subsection{Bag of Visual Words}
Once the features are extracted from each patch, 
we use the bag of words (BoW) representation  \cite{bow1} to describe each brain region, which calculates the histogram of representative patterns (or visual words) over all patches in this region.
To find the visual words, we apply the K-means clustering algorithm to the patch features obtained for all training patches. Given the MR images of a subject, we extract overlapping patches from each of the two designated brain regions (Thalamus and CC) Each patch is quantized to its closest visual words, and each metric in each region is described by a histogram of different visual words among all patches in this region. In the case where we trained a single auto-encoder to generate a single feature representation for all metrics together, each region is represented by a single histogram.
The block diagram of the BoW approach is shown in Figure 3.
In this figure, for better visualization we show patches from different parts of brain, but in our work we only focused on Thalamus and CC.
\begin{figure}[h]
\begin{center}
    \includegraphics [scale=0.27] {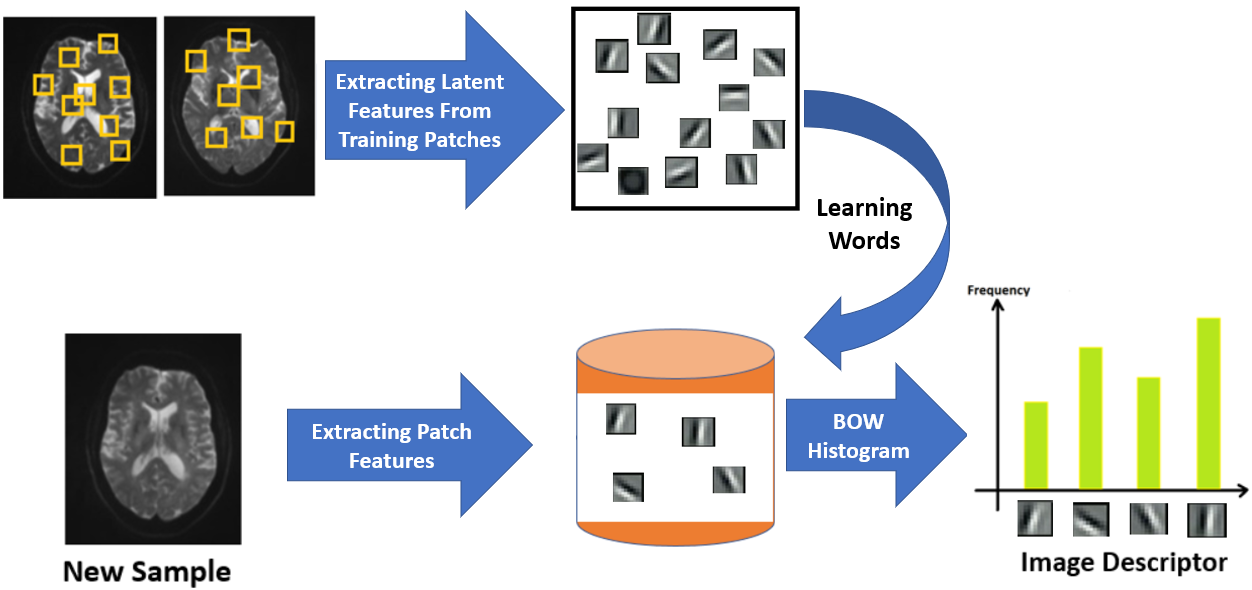}
\end{center}
  \caption{The block-diagram of the proposed BoW approach. }
\end{figure}

\subsection{Feature Selection and Classification}
After deriving deep-bag-of-words features from diffusion MR images, we will get two feature vectors, one for Thalamus region and the other for CC.
We concatenate the features from both regions (and all metrics), with demographic and neuro-cognitive test features, to form the final feature vector.
We then perform feature selection \cite{fs1}-\cite{fs2} to minimize the risk of over-fitting before classification.
We tried multiple feature selection approaches such as max-relevance and min-redundancy (MRMR) \cite{mrmr}, maximum correlation, greedy forward selection, and it turns out that the greedy forward feature selection works best for our problem.
This approach selects the best features one at a time  with a given classifier, through a cross-validation approach. 
For classification, we tried different classifiers (such as SVM, neural network, and random forest) and SVM was chosen because it generally gave better performance.

\section{Experimental Results}
We evaluate the performance of the proposed approach on our dataset of 114 subjects. 
This dataset contains 70 mTBI subjects between 18 and 64 years old, within 1 month of mTBI as defined by the American College of Rehabilitation Medicine (ACRM) criteria for head injury and 44 healthy age and sex-matched controls. 

To evaluate the model performance, we use  a cross validation approach, where each time we randomly take 20\% of the samples for validation, and the rest for training. 
We repeat this procedure 50 times (to decrease sampling bias), and report the average validation error as the model performance.

For the convolutional auto-encoder, we learn features on patches of 16x16 pixels in each image slice.
The encoder and decoder each have 4 layers, and the kernel size is always set to (3,3). 
The latent feature dimension is 32 for the networks which are trained on each metric, and 64 for the network which is trained on the stack of all metrics.
To train the model, the batch size is set to 500, and the model is trained for 10 epochs. The learning rate for the stochastic gradient descent is set to 0.0003.
The learnt auto-encoder is used to generate latent features on each overlapping patch in the training images. The resulting features are further clustered to $N$ words using K-means clustering. $N$ was varied among 20, 30, and 40. Each  MR metric in each region is represented by a histogram of $N$ dimension
For SVM, we use radial basis function (RBF) kernel. 
The hyper-parameters of SVM model (kernel width gamma, and the mis-classification penalty weight, C) are tuned based on the validation set.


In the first experiment, we compare the proposed approach with some previous works. 
For the proposed approach, the initial image feature representation is 260 dimensional, with 20 words for each of 8 MR metrics (AWF, DA, De\_par, FA, MD, AK, MK, RK) in CC, and 5 MR metrics in Thalamus (FA, MD, AK, MK, RK).
Together with additional 2 demographic and 4 clinical features (Stroop, SDMT, CVLT, FSS), the total feature dimension is 266.
We compare our approach with previous works that only used mean values of each MR metric in each region \cite{minaee1}, and also the work in \cite{yuanyi}, as well as BoW approach on raw patches \cite{minaee1} (that is, we find the visual words by applying the K-means clustering algorithm on the raw image patches directly).
It is worth  mentioning that the work in \cite{yuanyi} was using a different dataset, and an overlapping but different set of metrics.

With forward feature selection using the SVM classifier,  the  optimal feature subset contains 10 features FA in Thalamus, DA in CC, AWF in CC, AWF in CC, AWF in CC, FA in CC, RK in Thalamus, MK in Thalamus, RK in CC, FA in Thalamus. 
The comparison with previous works is provided in Table II.
As we can see, Deep-BoW approach achieves reasonable improvement over previously used features, except for BoW on raw patches where it performs slightly worse than using the raw pixel values. 
We believe the performance of Deep-BoW could be further improved by better tuning the deep network hyper-parameters.
\begin{table}[ht]
\centering
  \caption{Performance comparison of different approaches}
  \centering
\begin{tabular}{|m{5.3cm}|m{2.4cm}|}
\hline
The Algorithm  & Classification Rate on Validation Set\\
\hline 
Single best feature \cite{minaee1} &   \ \  \ \ \ \ \ \ \ 72\% \\
\hline
The selected subset with 8 features \cite{minaee1}  &  \ \ \ \  \ \ \  \ \ 80\%\\
\hline
The algorithm in \cite{yuanyi} using selected features with MRMR, and Baysian Network   &  \ \ \  \ \ \ \ \ \ 82\%\\
\hline
BoW on raw patches and 10 metrics \cite{minaee2}  &  \ \ \ \ \ \ \ \  \ 85.5\%\\
\hline
The algorithm in \cite{yuanyi} using selected features with MRMR, and Neural Network  &  \ \ \ \  \ \ \ \ \ 86\%\\
\hline
BoW on raw patches  with 20D histograms \cite{minaee1}  &  \ \ \ \ \ \  \ \   \ 92\%\\
\hline
The proposed Deep-BoW with 20D histograms  &  \ \  \ \  \ \ \ \ \ 87.8\%\\
\hline
The proposed Deep-BoW with 30D histograms  &  \ \ \  \  \ \ \ \ \ 90.1\%\\
\hline
\end{tabular}
\label{TblComp}
\end{table}


We have also evaluated the classification performance for feature subset of different size. 
Besides classification accuracy, we also report the sensitivity and specificity, which are important in the study of medical data analysis. 
The sensitivity and specificity are defined as in Eq. (1), where TP, FP, TN, and FN denote true positive, false positive, true negative, and false negative respectively. 
In our evaluation, we treat the mTBI subjects as positive. 
\begin{gather}
 \text{Sensitivity}= \frac{\text{TP}}{\text{TP+FN}} \ , 
\ \ \ \ \text{Specificity}= \frac{\text{TN}}{\text{TN+FP}} 
\end{gather}
Figure 4 denotes the classification accuracies, sensitivities and specificities achieved by optimum subset of feature of dimension 1 to 10.
\begin{figure}[h]
\begin{center}
    \includegraphics [scale=0.21] {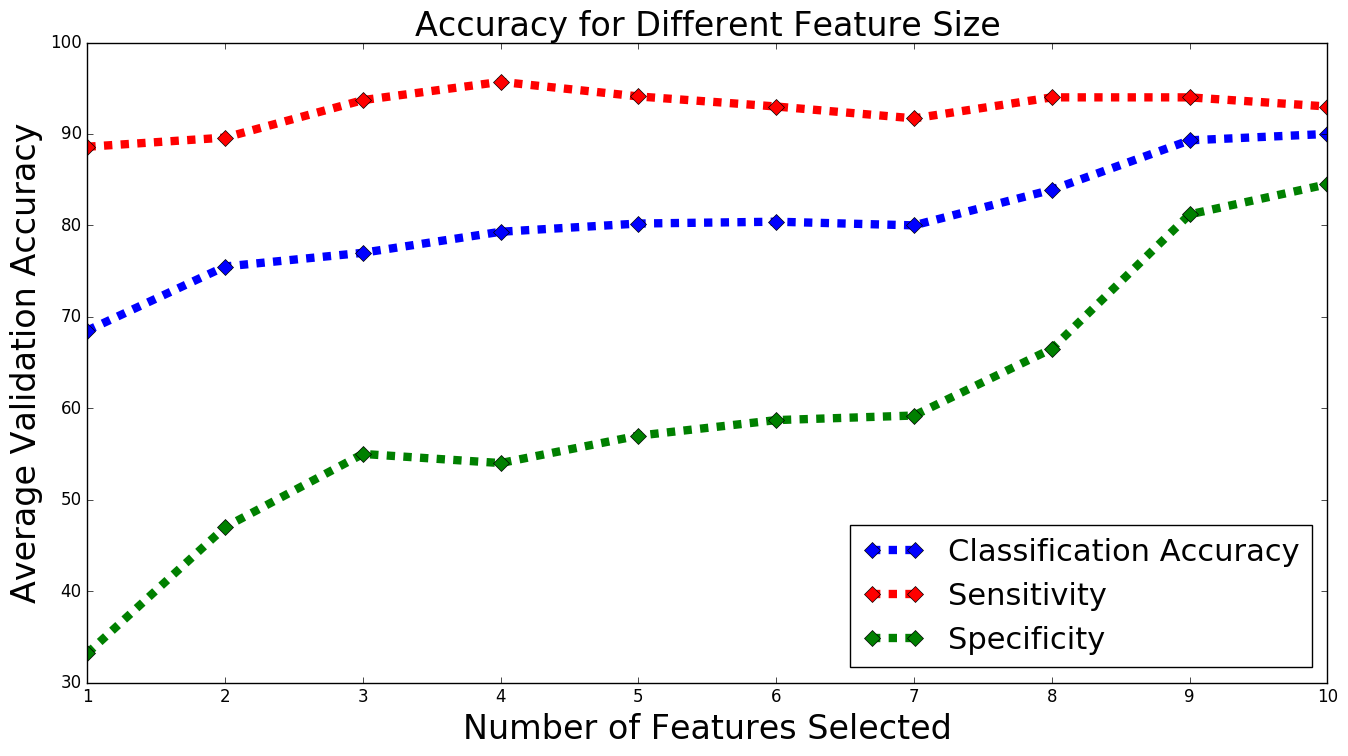}
\end{center}
\vspace{-0.2cm}
  \caption{The model performance for feature set of different size}
\end{figure}

In Table III, we provide a comparison between the two scenarios, where in one of them one network is trained per metrics, and in the other one a single network is trained over the stack of multiple metrics (for different number of words).

\begin{table}[ht]
\centering
  \caption{Performance comparison for different approaches}
  \centering
\begin{tabular}{|m{2.1cm}|m{1.25cm}|m{1.8cm}|m{1.5cm}|}
\hline
Auto-Encoder \ \ \ Scenario  & Bow-Hist Dimension & Image Feature Dimension & Cross-Validation Accuracy\\
\hline
Multiple Network  & 20 & 260 & 87.8\% \\
\hline
Multiple Network  & 30 & 390  & 90.1\%  \\
\hline
Single Network  & 130 & 260  & 89.4\% \\
\hline
Single Network  & 190 & 380  & 90\% \\
\hline
\end{tabular}
\label{TblComp}
\end{table}

To have a better generalization accuracy analysis, we also provide the comparison in terms of the accuracy on a heldout set.
In each run, we randomly pick 20 samples as the heldout set.
We then run cross validations 50 times within the remaining data, to generate 50 models, and use the ensemble of 50 models to make prediction on the held-out set and calculate the classification accuracy. 
We repeat this 6 times, each time with a different set of 20 heldout samples chosen randomly and report the average accuracy.
For this evaluation, we only compare two BoW based approaches, one derived by applying K-means on the raw patch images, and another by applying K-means on the patch level deep features. 
The average accuracy for the heldout sets for these two approaches are given in Table III. 
Although BoW using deep patch level features had lower cross validation accuracy in Table I, it shows equivalent heldout set accuracy.
For BoW on raw patches, the selected subset of features includes: FA in Thal (Thalamus), MK in Thal, FA in Thal,
Deperp in CC, MK in Thal, FA in CC, MD in
CC, AWF in CC, FA in CC.
For BoW using the deep features, the selected features include:  FA in Thal, DA in CC, AWF in CC, AWF in CC,
AWF in CC, FA in CC, RK in Thal, MK in Thal, RK in CC,
FA in Thal.
This comparison is provided in Table IV.
\vspace{-0.3cm}
\begin{table}[ht]
\centering
  \caption{Performance comparison of different approaches}
  \centering
\begin{tabular}{|m{4cm}|m{2.5cm}|}
\hline
The Algorithm   &  Heldout Set Accuracy \\
\hline
BoW on raw patches \cite{minaee1} using 20 words for each metric and region  &  \ \ \ \ \ \ \  \ 86.2\%\\
\hline
The proposed Deep-BoW  &  \ \ \ \ \ \ \ \  86.4\%\\
\hline
\end{tabular}
\label{TblComp}
\end{table}

Finally, we  present the the average histograms  of patients, and control subjects. 
These histograms and their difference are shown in Fig. 5.
As we can see mTBI and control subjects have clear differences in some part of these representations.
\begin{figure}[h]
\begin{center}
    \includegraphics [scale=0.18] {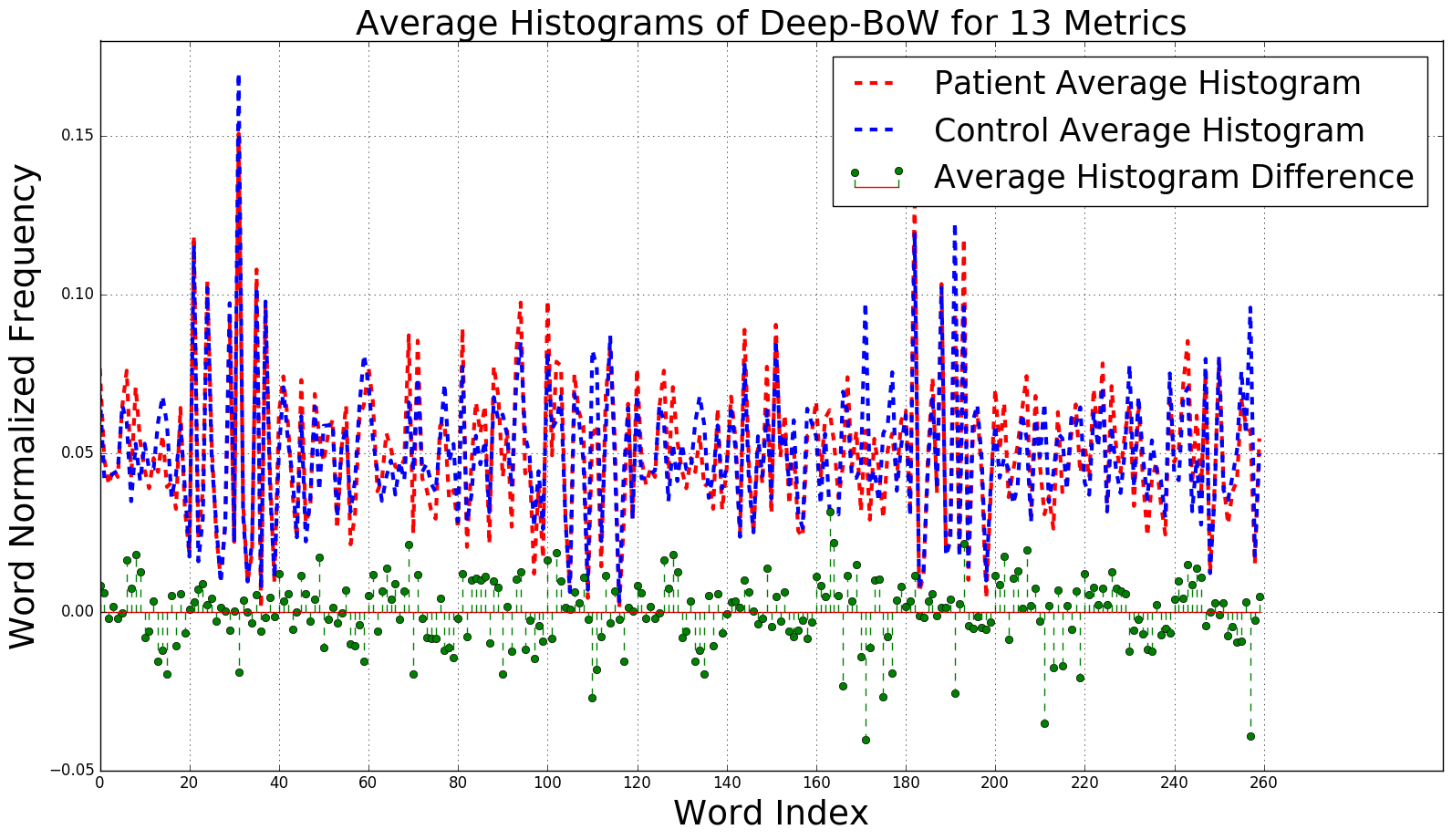}
\end{center}
\vspace{-0.2cm}
  \caption{Deep-BoW histograms of patients and controls}
\end{figure}

\section{Conclusion}
In this work, we propose a machine learning framework for mTBI identification from diffusion MRI using a relatively small dataset of 114 subjects.
To approach this problem without overfitting, we employ a deep unsupervised learning approach to learn feature representation for image patches, followed by aggregating patch level features using bag of word representation to form the overall image feature. 
These features are used along with demographic and neuro-cognitive features for patient identification.
Then greedy forward feature selection and SVM are used to perform classification.
Through experimental studies, we show that by learning deep visual features, we obtain significant gain over using mean values of MR metrics in brain regions, and also hand-crafted features.
 

\section*{ACKNOWLEDGMENT}
Research reported in this paper is supported by National Institute of Neurological Disorders and Stroke of  National Institutes of Health (NIH) under award number  R21NS090349. The content is solely the responsibility of the authors and does not necessarily represent the official views of the NIH.

\end{document}